\documentclass{article}
\usepackage{spconf,amsmath,graphicx}

\usepackage[nolist]{acronym}
\begin{acronym}[PHOENIX14\textbf{T} ]

\acrodefplural{rnn}[RNNs]{Recurrent Neural Networks}
\acrodefplural{cnn}[CNNs]{Convolutional Neural Networks}
\acrodefplural{hmm}[HMMs]{Hidden Markov Models}
\acrodefplural{gru}[GRUs]{Gated Recurrent Units}
\acrodefplural{crf}[CRFs]{Conditional Random Fields}
\acrodefplural{gan}[GANs]{Generative Adversarial Networks}
\acrodefplural{gpu}[GPUs]{Graphic Processing Units}

\acrodefplural{mdn}[MDNs]{Mixture Density Networks}

\acro{bpe}[BPE]{Byte Pair Encoding}
\acro{bsl}[BSL]{British Sign Language}
\acro{bleu}[BLEU]{Bilingual Evaluation Understudy}
\acro{blstm}[BLSTM]{Bidirectional Long Short-Term Memory}
\acro{bslcpt}[BSLCP\textbf{T}]{BSL Corpus \textbf{T}}
\acro{bobsl}[BOBSL]{BBC-Oxford British Sign Language}
\acro{cnn}[CNN]{Convolutional Neural Network}
\acro{crf}[CRF]{Conditional Random Field}
\acro{cslr}[CSLR]{Continuous Sign Language Recognition}
\acro{ctc}[CTC]{Connectionist Temporal Classification}
\acro{c4a}[C4A]{Content4All}
\acro{dl}[DL]{Deep Learning}
\acro{dgs}[DGS]{German Sign Language - Deutsche Gebärdensprache}
\acro{dsgs}[DSGS]{Swiss German Sign Language - Deutschschweizer Geb\"ardensprache}
\acro{dtw}[DTW]{Dynamic Time Warping}
\acro{fc}[FC]{Fully Connected}
\acro{ff}[FF]{Feed Forward}
\acro{gan}[GAN]{Generative Adversarial Network}
\acro{gpu}[GPU]{Graphics Processing Unit}
\acro{gru}[GRU]{Gated Recurrent Unit}
\acro{gtpt}[G2PT]{Gloss to Pose Transformer}
\acro{hmm}[HMM]{Hidden Markov Model}
\acro{hpe}[HPE]{Hand Pose Enhancer}
\acro{isl}[ISL]{Irish Sign Language}
\acro{lstm}[LSTM]{Long Short-Term Memory}
\acro{mha}[MHA]{Multi-Headed Attention}
\acro{mtc}[MTC]{Monocular Total Capture}
\acro{mse}[MSE]{Mean Squared Error}
\acro{mdn}[MDN]{Mixture Density Network}
\acro{mdgs}[MeineDGS]{Meine DGS-Annotated}
\acro{mdgst}[mDGS\textbf{T}]{meineDGS\textbf{T}}
\acro{mdgsth}[mDGS\textbf{T}-\textbf{H}]{meineDGS\textbf{T}-\textbf{HARD}}
\acro{mdgste}[mDGS\textbf{T}-\textbf{E}]{meineDGS\textbf{T}-\textbf{EASY}}
\acro{mt}[MT]{Machine Translation}

\acro{nmt}[NMT]{Neural Machine Translation}
\acro{nlp}[NLP]{Natural Language Processing}
\acro{ph12}[PHOENIX12]{RWTH-PHOENIX-Weather-2012}
\acro{ph14}[PHOENIX14]{RWTH-PHOENIX-Weather-2014}
\acro{ph14t}[PHOENIX14\textbf{T}]{RWTH-PHOENIX-Weather-2014\textbf{T}}
\acro{pof}[POF]{Part Orientation Field}
\acro{paf}[PAF]{Part Affinity Field}
\acro{pttt}[P2TT]{Pose to Text Transformer}
\acro{relu}[RELU]{Rectified Linear Units}
\acro{rnn}[RNN]{Recurrent Neural Network}
\acro{rouge}[ROUGE]{Recall-Oriented Understudy for Gisting Evaluation}
\acro{sgd}[SGD]{Stochastic Gradient Descent}
\acro{sla}[SLA]{Sign Language Assessment}
\acro{slr}[SLR]{Sign Language Recognition}
\acro{slt}[SLT]{Sign Language Translation}
\acro{slp}[SLP]{Sign Language Production}
\acro{smt}[SMT]{Statistical Machine Translation}

\acro{ttgt}[T2GT]{Text to Gloss Transformer}
\acro{ttpt}[T2PT]{Text to Pose Transformer}
\acro{ttp}[T2P]{Text to Pose}
\acro{ttg}[T2G]{Text to Gloss}
\acro{tth}[T2H]{Text to HamNoSys}
\acro{ttgth}[T2G2H]{Text to Gloss to HamNoSys}
\acro{ttgtp}[T2G2P]{Text to Gloss to Pose}
\acro{tts}[TTS]{Text to Speech}
\acro{wer}[WER]{Word Error Rate}

\end{acronym}

\usepackage{amsfonts}
\usepackage[capitalise]{cleveref}

\title{Gloss Alignment Using Word Embeddings}
\name{Harry Walsh, Ozge Mercanoglu Sincan, Ben Saunders, Richard Bowden
\thanks{We also thank the SNSF Sinergia project ‘SMILE II’ (CRSII5 193686) and the European Union’s Horizon2020 research project EASIER (101016982). This work reflects only the authors view and the Commission is not responsible for any use that may be made of the information it contains.}}
\address{CVSSP, University of Surrey\\ Guildford, United Kingdom \\
\{harry.walsh, o.mercanoglusincan, b.saunders, r.bowden\}@surrey.ac.uk}
\begin{document}
\ninept
\maketitle
\begin{abstract}
Capturing and annotating Sign language datasets is a time consuming and costly process. Current datasets are orders of magnitude too small to successfully train unconstrained \acf{slt} models. As a result, research has turned to TV broadcast content as a source of large-scale training data, consisting of both the sign language interpreter and the associated audio subtitle. However, lack of sign language annotation limits the usability of this data and has led to the development of automatic annotation techniques such as sign spotting. These spottings are aligned to the video rather than the subtitle, which often results in a misalignment between the subtitle and spotted signs. In this paper we propose a method for aligning spottings with their corresponding subtitles using large spoken language models. Using a single modality means our method is computationally inexpensive and can be utilized in conjunction with existing alignment techniques. We quantitatively demonstrate the effectiveness of our method on the \acf{mdgs} and \acf{bobsl} datasets, recovering up to a 33.22 BLEU-1 score in word alignment. 
\end{abstract}

\section{Introduction}

Sign languages are the primary form of communication for the Deaf. Signs are expressed through the articulation of manual and non-manual features including body language, facial expressions, mouthing, hand shape, and motion \cite{sutton1999linguistics}. 
Despite the recent successes of large language models, \acf{slt} between continuous sign language videos and spoken language remains a challenging task \cite{bragg2019sign}. Even though results have been achieved within a constrained setting and a limited vocabulary \cite{camgoz2018neural, camgoz2020sign}, progress towards unconstrained translation still requires larger-scale datasets. The visual nature of sign language has restricted the availability of high quality datasets, due to the difficulty of capturing and labeling a visual medium. The publicly available \ac{mdgs} dataset \cite{dgscorpus_3} attempts to fully capture the details of the language using gloss\footnote{Gloss is the written word associated with a sign} annotations, non-manual mouthing and the Hamburg Notation System (HamNoSys). However, the curation of such a high quality dataset is both time consuming and costly, which has restricted its size to only 50k parallel text gloss sequences \cite{dgscorpus_3}. This scarcity of data has motivated the research community to automate the collection and annotation of large-scale public datasets.

Broadcast content has repeatedly been used as the source of sign language datasets to assist with tasks such as sign language recognition, alignment, and translation \cite{Albanie2021bobsl, cooper2009learning, buehler2009learning}. Under the European Accessibility Act, all EU countries are obligated to make content accessible \cite{eu03}. Specifically, UK broadcasters must supply 5\% of their content with \ac{bsl} translations, which leads to the generation of a steady stream of sign language translation data. However, the raw data only contains the spoken language subtitles and the video of the sign interpreter, who, although conducting translations from the subtitles, is often misaligned. In order to make use of this data for tasks such as \ac{slt}, the data needs to be curated, and subsequently aligned. 

\begin{figure*}[htbp!] 
\begin{center}
\includegraphics[width=0.98\textwidth]{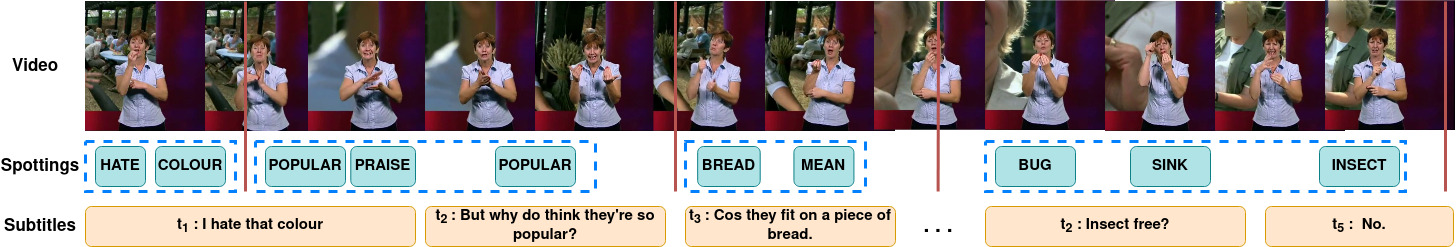}
\caption{A visualisation of the alignment from the \acf{bobsl} dataset (vertical red lines indicate the correct boundaries).}
\label{fig:vis_alignment}
\end{center}
\end{figure*}

As shown in \cref{fig:vis_alignment}, we have identified two types of alignment error; 1) Glosses that correspond to the preceding sentence are aligned to the current, shown by the gloss POPULAR and PRAISE that are misaligned to sentence \(t_{1}\). 2) Glosses are aligned to the following sentence, shown by the gloss INSECT that is misaligned to \(t_{5}\). There are several factors that lead to the misalignment of the sign to the spoken language subtitle. Firstly, there is a weak correlation between the number of words in a sentence and the number of signs contained in the translation. Additionally, the time taken to speak a word is not related to the time taken to perform a sign. Finally, the ordering of spoken language words is different from the gloss order \cite{sutton1999linguistics}. All these factors result in the sign language lagging or preceding the corresponding subtitle.

Previous work has attempted to align the subtitles with the sign language video by finding a correspondence between the glosses of the spotted isolated signs and words in the subtitle with a similar lexical form  \cite{albanie2020bsl, momeni2020watch, varol2021read}. However, these works all require multi-modal inputs and are expensive to compute. 

In this paper, we propose an approach to align glosses to the corresponding spoken language sentence by leveraging the power of large spoken language models, such as BERT \cite{devlin2018bert} and Word2Vec \cite{mikolov2013distributed}. We make the following 2 contributions: 1) A novel alignment approach that can be used in conjunction with previous methods. 2) Quantitative evaluation of our approach on two different datasets from both \ac{bsl} and \ac{dgs}, demonstrating our approach is language agnostic.

\section{Related Work}

\subsection{Spoken Language Alignment}

Word alignment techniques have been researched since the late $20^{\text{th}}$ century, where Brown et al. developed the IBM models to assist in statistical machine translation \cite{brown1992class}. Since then a number of statistical word alignment techniques have been proposed, such as GIZA++ \cite{och2000improved, gao2008parallel} or alignment via a hidden Markov model \cite{vogel1996hmm}. More recently, deep learning based methods have demonstrated superior performance \cite{stengel2019discriminative}. A variety of supervised methods have been created, some using statistical supervision \cite{tamura2014recurrent}, while others make use of the attention mechanism from a transformer \cite{zenkel2019adding}. Most similar to our approach, Stengel-Eskin et al. used the dot product distance between learnt embeddings to make an alignment prediction \cite{stengel2019discriminative}. 

\subsection{Sign Language Spotting}

Sign spotting is the task of locating isolated instances of signs in a continuous video. Several methods have been suggested to tackle this task, from early techniques that use hand crafted features \cite{yang2008sign, santemiz2009automatic}, to methods that employ subtitles as a form of weak supervision \cite{cooper2009learning, buehler2009learning}. More recent methods have employed multiple modalities to improve performance e.g. visual dictionaries \cite{momeni2020watch} and mouthings \cite{albanie2020bsl}. However, all these methods still result in the misalignment of spotted signs and the subtitles, as shown in \cref{fig:vis_alignment}.  

\subsection{Subtitle Alignment}

Subtitle alignment attempts to align a continuous sequence of signing to the corresponding subtitles. Early attempts to solve the alignment issues used 3D pose in a multi step approach, but assumed a similar ordering between the spoken and signed languages \cite{farag2019learning}. To overcome this assumption, Bull et al. trained a \ac{blstm} with 2D keypoints using manually aligned subtitles as ground truth to segment a continuous video into candidate signs \cite{bull2020automatic}. However, without a strong language model, such approaches tend to over segment the video. In subsequent works, the subtitles were incorporated into the input of the model along with the video and shifted temporal boundaries, to align broadcast footage \cite{bull2021aligning}. In contrast, in this work we attempt to align spotted glosses to the spoken language subtitles using only word embeddings. Note our approach can be used in conjunction with these existing methods.

\section{Methodology}

In this section we explain our methodology for aligning glosses with their corresponding subtitles. In \cref{sec:mapping} we explain how we use the embeddings from large spoken language models such as BERT and Word2Vec to create a mapping between a sequence of glosses and spoken language words. Then in \cref{sec:gloss_alignment} we show how to use the mapping to re-align glosses to the correct spoken language sentence. 

\subsection{Text Gloss Mapping} \label{sec:mapping}

Our alignment approach relies on the lexical overlap that exists between the spoken language words and the signed glosses. Therefore, the gloss notation needs to be semantically motivated. Given the following example text, ``where do you live?'' and the following sequence of glosses, ``YOU LIVE WHERE. ME LONDON''. It is clear to see that the first three glosses correspond to the given text and the last two glosses potentially correspond to the next sentence.  

Following this intuition, we use two different word embedding techniques to find which glosses best correspond to a given spoken language sentence. Firstly, we use Word2Vec \cite{devlin2018bert} to find connections between words and glosses that have a similar lexical form. Secondly, we use BERT \cite{mikolov2013distributed} to find connections based on meaning. We find BERT embeddings capture the meaning of words allowing us to find connections between words and glosses that have a different lexical form, e.g. "supermarket" and "SHOP". Note when we apply our approach to \ac{dgs} we first apply a compound splitting algorithm to improve the performance \cite{tuggener2016incremental}.

\begin{figure}[!ht]
    \centering
    \includegraphics[width=0.45\textwidth]{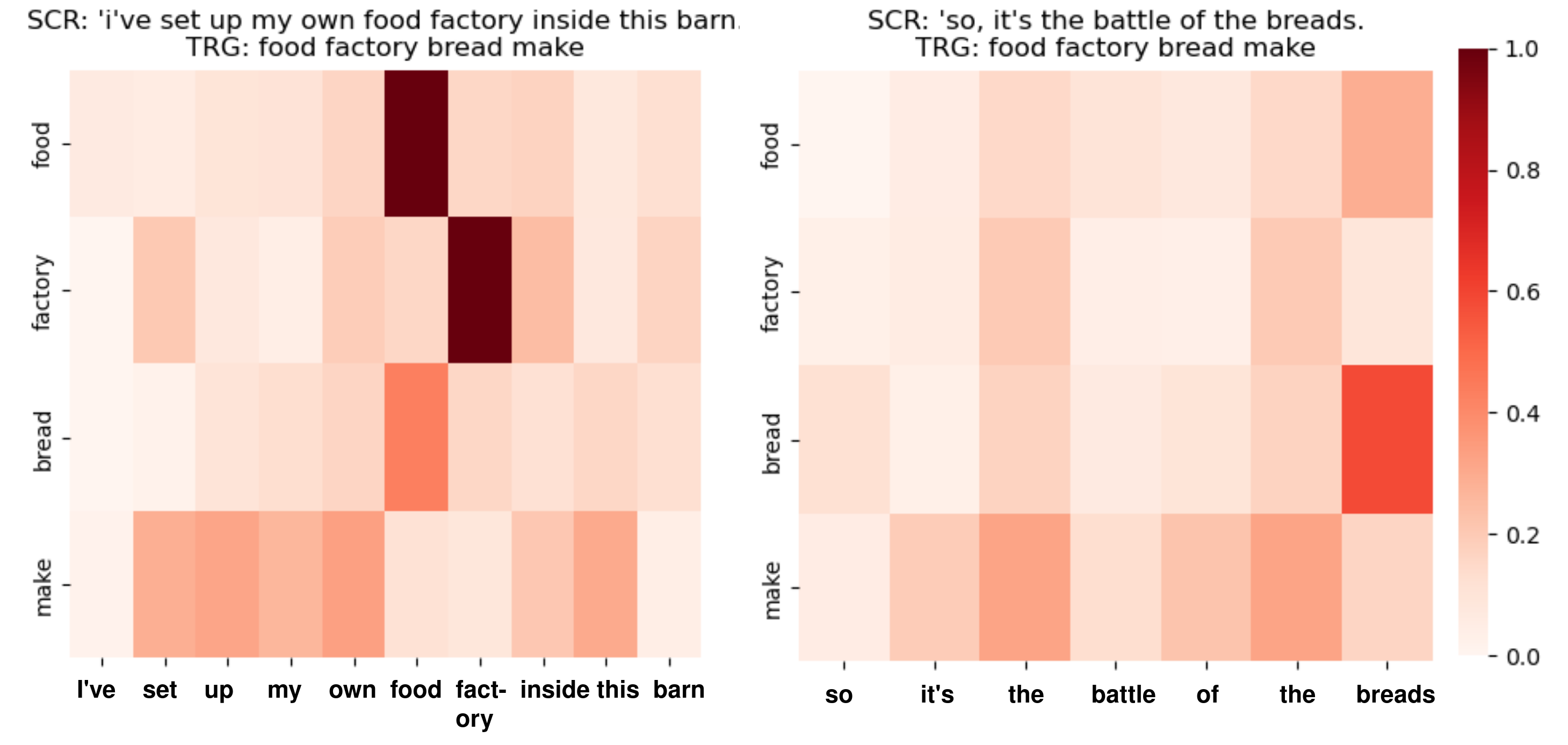}
    \caption{An example of two alignments, \(A\), found between the TRG: ``FOOD FACTORY BREAD MAKE'' and sentence one (left): ``I've set up my own food factory inside this barn'', and sentence two (right): ``So, it's the battle of the breads.''.}
    \label{fig:bobsl_alignement_heatmap_12}
\end{figure}

To find a mapping between a spoken language sequence \(X = (x_{1},x_{2},...,x_{W})\) with W words, and a sequence of glosses, \(Y = (y_{1}, y_{2},...,y_{G})\) with G glosses, we first apply Word2Vec; 
\begin{equation} X_{Vec} = Word2Vec(X) \end{equation}
\begin{equation} Y_{Vec} = Word2Vec(Y) \end{equation}
where \( X_{Vec} \in \mathbb{R}^{W \times E_{vec}}\) and \( Y_{Vec} \in \mathbb{R}^{G \times E_{vec}}\). Calculating the outer product between the two embeddings produces the Word2Vec alignment;
\begin{equation} A_{Vec} = Y_{Vec} \otimes X_{Vec} \end{equation}
where $ A_{Vec} \in \mathbb{R}^{G \times W}$. We repeat the above using BERT, to find connections based on meaning; 
\begin{equation} X_{BERT} = BERT(X) \end{equation}
\begin{equation} Y_{BERT} = BERT(Y) \end{equation}
\begin{equation} A_{BERT} = Y_{BERT} \otimes X_{BERT} \end{equation}
where $ X_{BERT} \in \mathbb{R}^{W \times E_{BERT}}$, $ Y_{BERT} \in \mathbb{R}^{G \times E_{BERT}}$ and $ A_{BERT} \in \mathbb{R}^{G \times W}$. The BERT model we apply uses a wordpiece tokenizer, therefore to find an alignment on the word level we average the embeddings of the sub-units. 
The final alignment is found by joining the alignments from BERT and Word2Vec. We filter the Word2Vec alignment scores by \(\alpha\), only keeping strong connections. Thus, the final alignment is defined as; 
\begin{equation} Align(X, Y) = A_{BERT} + (\alpha * A_{Vec}) \end{equation}

\noindent where $ A \in \mathbb{R}^{G \times W}$. A visualization of two alignments, \(A\), is shown in \cref{fig:bobsl_alignement_heatmap_12}. Here we find the alignment between two sequential text sentences from the \ac{bobsl} dataset, and the four glosses that correspond to the two sentences. FOOD and FACTORY belong to the first sentence "I've set up my own food factory inside this barn", and as shown by \cref{fig:bobsl_alignement_heatmap_12} (left) a strong alignment is found between the spoken language and it's corresponding glosses, FOOD, and FACTORY. The same applies to \cref{fig:bobsl_alignement_heatmap_12} (right) with the gloss BREAD.  

\subsection{Gloss Alignment} \label{sec:gloss_alignment}

We use the alignment found above for two sequential sentences to re-align glosses to their corresponding subtitles.  Given a dataset that consists of N sequences of spoken language sentences \(T = (X_{1},X_{2},...,X_{N})\) and N sequences of glosses \(S = (Y_{1},Y_{2},...,Y_{N})\), we take two sequential sentences, \(X_{i},X_{i+1}\), and we concatenate their corresponding glosses \(Y_{i:i+1} = Y_{i} + Y_{i+1}\). 

\begin{figure}[!ht]
    \centering
    \includegraphics[width=0.4\textwidth]{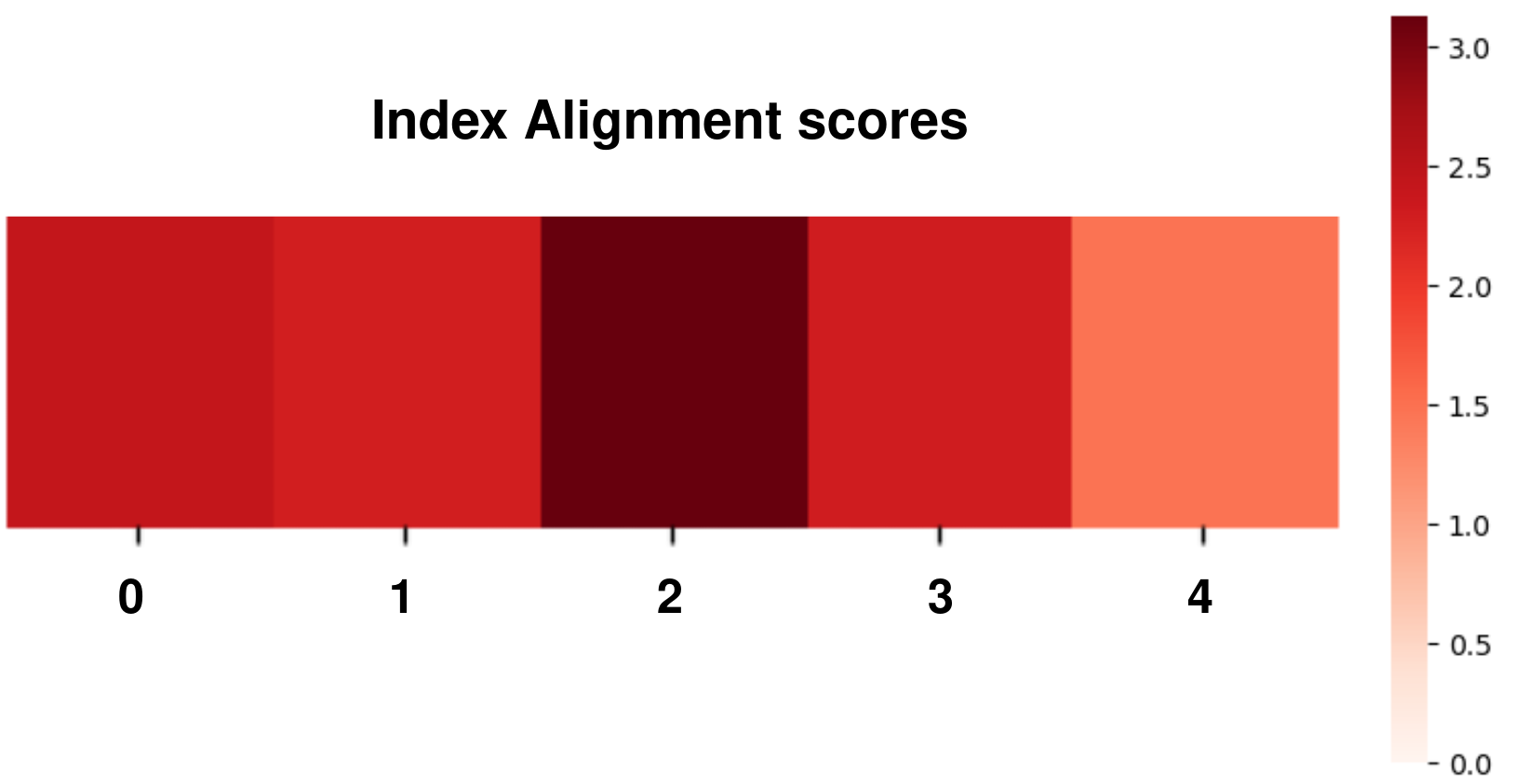}
    \caption{The alignment scores found for the two sentences in \cref{fig:bobsl_alignement_heatmap_12}}
    \label{fig:alignment_score}
\end{figure}

We want to find the index to split \(Y_{i:i+1}\) back into two sequences that result in the best alignment. The number of possible splits is \(N_{split} = length(Y_{i:i+1}) + 1 \). Therefore, for each possible split we sum the max alignment score for each gloss in \(Align(X_{i}, Y_{i:i+1})\) and \(Align(X_{i+1}, Y_{i:i+1})\) as in Equation 7. \cref{fig:alignment_score} shows the alignment score for each possible split of \(Y_{i:i+1}\). We take the argmax of the alignment score to determine the optimal index to split \(Y_{i:i+1}\) back to two sequences.

\cref{fig:alignment_score} shows the alignment score for the two sentences in \cref{fig:bobsl_alignement_heatmap_12}, showing our approach is able to find the optimal alignment. 

The proposed algorithm is auto-regressive, meaning the output of the first split affects the next iteration. This introduces a bias that favours earlier sentences in the dataset. Therefore, to counter this effect we iterate through the data from \(i = [0, 1, 2, ... N]\) and then for each subsequent iteration we reverse the order, such that iteration two is \(i = [N, N-1, N-2, ..., 0]\). In the next section we show that multiple iterations (forwards then backwards) of the algorithm increase the alignment score, but quickly converges.

\section{Experimental Setup}
In this section we outline the experimental setup, detailing the pre-trained models that we use to create the word embeddings for both English and German. In \cref{sec:method_mdgs} we describe how we corrupt the \ac{mdgs} dataset to simulate a spotting misalignment. Finally, in \cref{sec:method_bobsl} we explain how we gather the spottings from \cite{bull2021aligning} and process them to create parallel text gloss sequences for our alignment algorithm. 

We use the Fasttext implementation of Word2Vec that supports 157 languages \cite{grave2018learning}. The models are trained on the Common Crawl and Wikipedia datasets and have an output dimension of 300. For the following experiments we use the English implementation when testing our approach on the \ac{bobsl} dataset and the German version when testing on \ac{mdgs}. Note, we filter the Word2Vec embeddings by setting \(\alpha\) to 0.9.  

When creating embeddings with BERT we use Huggingface's python library transformers to load the models. When testing on \ac{mdgs} we use Deepsets implementation of German BERT \cite{chan-etal-2020-germans}, which is trained on approximately 12GB of data from the Wiki, OpenLegalData, and News datasets. Finally, when testing on the English \ac{bobsl} dataset we use GoogleAI's implementation of BERT \cite{DBLP:journals/corr/abs-1810-04805}, which is trained on the Bookcorpus and Wikipedia datasets. To evaluate the performance of our algorithm on all datasets we use BLEU-1 score. We do not present results using higher n-gram BLEU scores as these metrics are used to measure the order accuracy, that is unnecessary for this task. 

\subsection{\ac{mdgs} Dataset} \label{sec:method_mdgs}

All results on the \ac{mdgs} dataset are computed against the original ground truth. The \ac{mdgs} dataset contains 50k parallel sequences \cite{dgscorpus_3} and we follow the translation protocol set in \cite{saunders2021signing}. The dataset has a source vocabulary of 18,457 with 330 deaf participants performing free form signing. Note we reorder the sequences sequentially as in the original videos.  

To evaluate our approach we corrupt the \ac{mdgs} dataset. This allows us to simulate an alignment error created when using previously mentioned sign spotting techniques to automatically spot glosses in a sequence of continuous signing. We create two versions of the dataset to simulate; 

\subsubsection{Sequence misalignment} 
A worst-case scenario, a total misalignment of all sequence pairs. For this, we offset all the gloss sequences by one. We add an empty sequence to the start \(Y_{empty}\) and remove the last sequence \(Y_{N}\) to maintain an equal number, N, of text gloss pairs. Therefore, we apply our alignment approach to \(T = (X_{1},X_{2},...,X_{N})\) and \(S = (Y_{empty},Y_{1},...,Y_{N-1})\). 

\subsubsection{Gloss misalignment} 
To simulate the errors shown in \cref{fig:vis_alignment} (glosses are misaligned to the preceding or succeeding sentence) we randomly shift up to 3 glosses to the previous or following sequence. We set probabilities of 15\%, 20\% and 10\% of moving 1, 2 or 3 glosses, respectively. 10\% of the time we do not alter the sequence. Note if the sequence has fewer glosses than we wish to shift, then we do not alter it. In total we move 21,273 glosses to the preceding sequence and 21,359 to the next sequence. 

\subsection{\ac{bobsl} Dataset} \label{sec:method_bobsl}

The \ac{bobsl} dataset contains 1,193K sentences extracted from 1,962 videos from 426 different TV shows \cite{Albanie2021bobsl}. The dataset itself only contains the subtitle from the original TV show and the signer. The test set comes with three variants of the subtitles audio-aligned, audio-aligned shifted, and manually aligned. By calculating the average difference between the audio-aligned and signing-aligned sentences, Albanie et al. \cite{Albanie2021bobsl} found the signer lags the subtitle by approximately +2.7 seconds. Thus, the audio-shifted variant applies this 2.7 second delay to all time stamps. The manually aligned subtitles contain a subset of the original audio-aligned subtitles. Therefore, when comparing our alignment results against the manually aligned subtitles we are restricted to this subset of the data.

In order to perform alignment we use the automatically extracted spottings from \cite{momeni2022automatic}. We then process the data, aligning the dense spots with the three variants of the subtitle provided in the original \ac{bobsl} test set.

\section{Experiments - Quantitative Evaluation} 

\begin{figure}[!ht]
    \centering
    \includegraphics[width=0.45\textwidth]{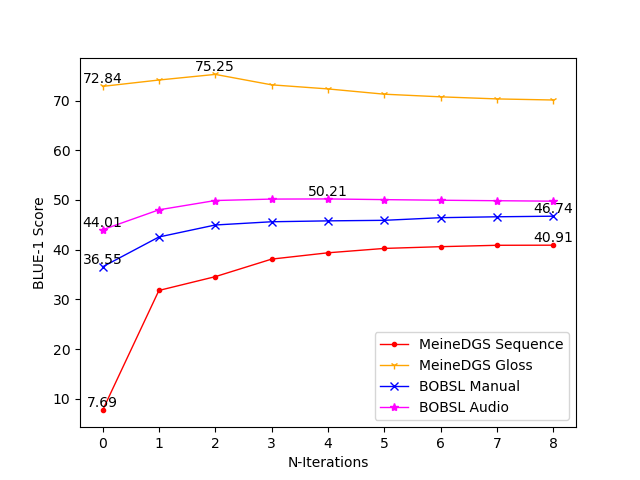}
    \caption{Results of applying the alignment algorithm to the \ac{mdgs} (Gloss and Sequence level misalignment) and \ac{bobsl} datasets (Audio aligned and Manually aligned)}
    \label{fig:alignment_results}
\end{figure}

\cref{fig:alignment_results} shows the results of applying the alignment algorithm to two versions of the \ac{mdgs} and \ac{bobsl} datasets. As can be seen, the algorithm has a positive effect on all variants, increasing the BLEU-1 scores by up to 33.22. Next we discuss the results in detail, starting with \ac{mdgs} (sequence level misalignment, then gloss level misalignment), followed by the \ac{bobsl} dataset (audio aligned, then manually aligned). 

\subsection{\ac{mdgs} Alignment}

\subsubsection{Sequence misalignment} 
In this experiment we offset the \ac{mdgs} dataset by 1 sequence, to simulate the worst case where all glosses are misaligned. \cref{fig:alignment_results} - \ac{mdgs} Sequence (orange line) shows there is a shared gloss vocabulary between sequential sentences, as the baseline score is not zero. Impressively, the approach is able to recover a large proportion of the glosses, increasing the BLEU-1 score from 7.69 to 40.91, a improvement of 432\%.

\subsubsection{Gloss misalignment} 
\cref{fig:alignment_results} - \ac{mdgs} Gloss (yellow line) shows the results of applying our alignment approach to the corrupted dataset. By corrupting the data we decrease the BLEU-1 score from perfect alignment (100 BLEU-1) to 72.84. From this baseline we are able to recover 2.41 BLEU-1 score using a single forward and backward pass through the data, an improvement of 3.3\%. However, further iterations are detrimental as can be expected from a greedy algorithm. 

This shows that the approach is able to recover a portion of the corruption. However, it should be noted that the effectiveness of the approach is dataset dependent, as the similarity of sequential sentences will effect the reliability of the mapping found between words and glosses. 

\subsection{\ac{bobsl} Alignment}

\subsubsection{Audio aligned} 
Here we show that our approach is able to move the audio-aligned subtitle toward the improved audio-shifted subtitles. As shown in \cref{fig:alignment_results} the approach improves the audio alignment by 6.2 BLEU-1. It should be noted that the audio-shifted subtitles are not perfect ground truth. Additionally, the spottings are not perfect, which introduces an error to any alignment approach as we may be attempting to map glosses that do not align with any words in the spoken sentence. Thus, we could expect the performance to increase if the quality of the underlying spottings improves. 

\subsubsection{Manually aligned} 
The manually aligned subtitles and their timings affect how we collect the spottings, which leads to a variation in the number of glosses. Hence, why the baseline score at \(N = 0\) is lower compared to the previous audio-aligned experiment. Despite this limitation, the approach is able to improve the alignment by 10.19 BLEU-1.

\section{Conclusion}
Sign Language alignment is an essential step in creating large-scale datasets from raw broadcast data. Improving the alignment between the subtitles and the associated signed translation would have positive effects on tasks such as translation, recognition, and production. In this paper we have demonstrated that embeddings from large spoken language models can be used to align glosses with their corresponding subtitles. Our approach can be run in addition to existing multi-model methods and is computationally inexpensive in comparison. We have shown the approach is capable of recovering up to a 33.22 BLEU-1 score in word alignment. 

\section{ACKNOWLEDGMENT}
We thank Adam Munder, Mariam Rahmani, and Marina Lovell from OmniBridge, an Intel Venture, for supporting this project. We also thank Thomas Hanke and the University of Hamburg for use of the \ac{mdgs} data.

\vfill\pagebreak

\bibliographystyle{IEEEbib}
\bibliography{refs}

\end{document}